# Automated Utterance Labeling of Conversations Using Natural Language Processing

Maria Laricheva[1][0000-0003-0369-387X], Chiyu Zhang[1], Yan Liu[2], Guanyu Chen[1], Terence Tracey[1,3], Richard Young[1], Giuseppe Carenini[1]

[1] The University of British Columbia, Vancouver BC, Canada
[2] Carleton University, Ottawa ON, Canada
[3] Arizona State University, Tempe AZ, USA
`maria.laricheva@ubc.ca`

**Abstract.** Conversational data is essential in psychology because it can help researchers understand individuals' cognitive processes, emotions, and behaviors. Utterance labelling is a common strategy for analyzing this type of data. The development of NLP algorithms allows researchers to automate this task. However, psychological conversational data present some challenges to NLP researchers, including multilabel classification, a large number of classes, and limited available data. This study explored how automated labels generated by NLP methods are comparable to human labels in the context of conversations on adulthood transition. We proposed strategies to handle three common challenges raised in psychological studies. Our findings showed that the deep learning method with domain adaptation (RoBERTa-CON) outperformed all other machine learning methods; and the hierarchical labelling system that we proposed was shown to help researchers strategically analyze conversational data. Our Python code and NLP model are available at https://github.com/mlaricheva/automated_labeling.

## 1 Introduction

Analyzing conversational data is essential in psychology because it can help researchers understand individuals' cognitive processes, emotions, etc. One of the conventional methods is to code/label the conversations line by line, also called utterances, by two independent researchers based on a pre-established coding list, which enables further qualitative data analysis. However, the coding is a time-consuming and labor-intensive process that takes approximately 4-6 times of the video recording time [1].

The recent development in natural language processing (NLP) makes it possible to automatically label utterances and significantly reduce the coding time. Some researchers have explored automated labeling for counselling data. For example, Tanana et al. [2] and Lee et al. [3] applied dialogue act classification to therapy transcripts to understand the flow of conversation in therapy sessions. Can et al. [1] conducted automated labeling of counselor's reflections. However, most of these studies do not extensively address the existing challenges of the psychological data.

**One challenge** is multilabel classification. In psychological conversational data, the behaviours and emotions are often co-occurring, i.e., each utterance may have more than one label. Multilabel classification is not naturally supported by conventional machine learning algorithms, such as Naïve Bayes. **Another common challenge** is a detailed coding system that includes many labels. A large number of classes leads to data imbalance and confounds the classification problem. **The third challenge** is a limited amount of the training data. Due to the privacy concerns and annotation costs, psychological data is hard to access and collect [1, 4], which aggravates the first two challenges.

Several strategies exist to manage mentioned problems. To deal with the multilabel classification (Challenge 1), a task can be decomposed into a series of binary classification tasks using the classifier chain [5]. In the case of a large number of labels (Challenge 2), researchers attempt to aggregate the fine-grained classes into more coarse-grained categories [6-8]. The lack of relevant training data (Challenge 3) may be solved using unsupervised learning algorithms, but in that case, the resulting classes are hard to explain. Another approach to solving the data scarcity is transfer learning from the rich-resource task to a new low-resource task to improve the performance of the latter [9]. A few studies on counseling applications of NLP paid attention to these data problems and none solved them simultaneously.

This study is to investigate how automated labels generated by NLP methods are comparable to human labels in the context of conversations on adulthood transition. We compared several NLP methods, including conventional machine learning and deep learning methods. More specifically, we provided strategies to handle three challenges discussed earlier: (1) demonstrated how to resolve the challenge of one utterance having multiple labels; (2) introduced a hierarchical labeling system to strategically analyze utterances, generate automated labels from general categories to more refined labels; and (3) explored how to pretrain a deep learning model, RoBERTa-CON, on a large data set of counselling conversational data.

## 2 Data

### 2.1 Source

We used the data that was collected by Young et al. [10-11] and that focuses on youth adulthood transition between 2013 and 2016. We utilized a dataset that consisted of 63 text transcripts, each corresponding to the session with a duration of around 10-20 minutes. The data included conversations between peers (25 transcripts), parents or a parent and an older sibling (29 transcripts), and a parent and a child (9 transcripts). Participants discussed topics that they considered relevant to the theme of the transition to adulthood.

The coding scheme for the original dataset was developed with a perspective of an action theory that conceptualizes the conversation as a joint goal-directed action [12]. After the manual data cleaning, we decided to merge similar labels (e.g., labels '*states*

*opinion'* and *'states perception'* were combined into a joint class *'express perception or opinion'*), and the resulting coding system comprised of 59 unique labels.

## 2.2 Preprocessing

The final dataset included 7,965 utterances. The data were split into train (90%) and test (10%) sets using stratified sampling to ensure the inclusion of underrepresented labels (such as the label praise with only 12 utterances, which is less than 0.007% of the emotion dataset).

Context is essential for a complete utterance-level dialogue understanding. For our dataset, we extended each utterance with the previous two: to predict the label $y_i$ of the utterance $u_i$, we used the combination of $u_{i-2}$, $u_{i-1}$ and $u_i$ utterances. The placeholder was used if there were no utterances preceding.

For conventional machine learning methods, utterances were cleaned by removal of stop-words, punctuation and special symbols and then transformed to numerical vectors, using term-frequency inverse-document-frequency (TF-IDF) technique [13]. For deep learning methods (e.g., RoBERTa), we used the original text without any modification to train the model end-to-end.

## 2.3 Challenges

**Multilabel**. A multilabel classification implies that more than one label can be assigned to each utterance. In the original dataset, about 24% of utterances were labeled by two classes, and 12% by three or more. See an example in Figure 1.

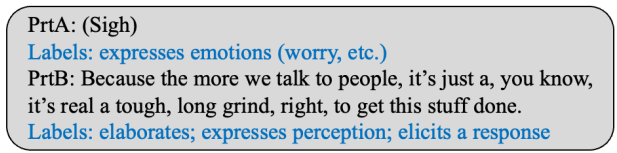

**Fig. 1.** An example of a snippet of conversation between a couple

**Multiclass.** As was mentioned previously, the psychological data has a challenge of large number of labels. To tackle this problem, we introduce the hierarchical labeling system. At the top level, the initial labels constituted two large categories: expressions of emotions and cognitive processes and behaviors. This dichotomy was preserved in our analysis. We refer to this task as **EMO-COG**. The corresponding analysis was conducted on the whole dataset, consisting of 7,965 utterances, with 1,373 utterances corresponding to emotion and 6,592 to non-emotion.

For each of the main categories, we developed an independent higher-level classification consisting of eight classes. For 1,373 emotional utterances, the higher-level classification was based on but not identical to the fundamental emotions from Plutchik's model [14]. We linked 20 original labels to eight classes and referred to this classification as **EMO-8**. For 6,592 utterances corresponding to cognitive processes and behaviors, we developed a coding system by grouping together similar labels. For example, the higher-level category *'clarification'* included the following original labels: *'paraphrase'*, *'ask for clarification'*, *'ask for confirmation'*, *'clarify'*,

'*confirm*', '*elaborate*'. The resulting task included eight cognition labels instead of the original 39 and is referred to as **COG-8**.

Finally, original labels for emotions and cognitive processes and behaviors independently constituted two fine-grained categorization tasks. The fine-grained emotion classification consisted of 20 unique categories and was referred to as **EMO-FULL**. The fine-grained classification for cognitive processes included 39 unique classes and was referred to as **COG-FULL**.

## 3 Methods

### 3.1 Conventional Machine Learning Methods

We utilized five popular conventional machine learning methods [13] in the data analysis: (1) Naïve Bayes classifier (**NB**), a probabilistic algorithm, has been widely used for the multiclass text classification; (2) AdaBoost is a boosting ensemble method used to improve the model's performance. We used AdaBoost to enhance the base Naïve Bayes (**AdaBoost+NB**); (3) Random Forest (**RF**) is another ensemble learning method, which can prevent the model from overfitting; (4) Gradient Descent (**GD**) is a popular optimization technique that is used together with linear models. We used support vector machine (**SVM**) as a backbone for the GD; and (5) Logistic regression (**LR**) algorithm applied with L2-norm penalty.

For EMO-COG, we conducted binary classification analysis with all the models described above. For multilabel tasks, we trained each model within a classifier chain framework in order to predict multiple labels for a single utterance.

### 3.2 Deep Learning Methods

Recently, deep learning models have revolutionized the research in NLP and have been shown superiority to conventional machine learning methods [15]. The deep learning models are data-hungry, i.e., require a large amount of training data. However, our data is limited to train the deep learning model. Hence, we proposed to solve this challenge by using the state-of-the-art pretrained language model.

In our study, we used **RoBERTa** (Robustly Optimized BERT Pretraining Approach [16]) which was trained on a large English dataset, such as Wikipedia and BookCorpus. We utilized the base architecture of RoBERTa that includes 12 Transformer encoder layers, 12 heads each, and 768 hidden units [17]. We used hidden states of "[CLS]" token from the last layer of RoBERTa as a sequence-level representation vector of each input utterance. For EMO-COG, we sent this vector through a non-linear layer with softmax activation function to predict the label and trained the model with cross-entropy loss. For multilabel classifiers, we adopted a similar approach, but we used a sigmoid activation function and trained with binary cross-entropy loss (BCE).

**Domain Adaptation**. For deep learning models, the performance deteriorates significantly when there is a domain shift [18]. As was mentioned previously, RoBERTa models were pretrained on data that is different from our domain (psychological conversations), therefore, we decided to perform a domain adaptation to improve the model's performance on our tasks. The domain adaptation is a technique that uses a larger dataset of a similar domain, closer to the target, to pre-train the model and learn the relevant characteristics of the target data. In this experiment, we exploited the Alexander Street Press database on Counseling and Psychotherapy Transcripts [19]. This dataset consisted of more than 1,500 conversations between a therapist and a patient. Therefore, we further pretrained RoBERTa on this dataset with masked language modeling objectives [16] which mask 15% of input tokens and task the model to reconstruct the original input sequence. The final model is referred to as **RoBERTa-CON**.

### 3.3 Hyperparameter

For **conventional machine learning** methods, we used the TF-IDF matrix with a vocabulary size of 3,034. For the AdaBoost + NB, we set the number of estimators as 50, and the learning rate as 0.1. For RF, we used the default hyperparameters of sci-kit learn implementation, which sets the number of estimators as 100 and uses the split criterion of Gini [17]. GD-SVM model was trained with 1,000 epochs and the learning rate was identified by the 'optimal' mode of sci-kit learn. For LR, we used L2-norm penalty and trained it for up to 100 iterations. All conventional models were trained on the training dataset and tested on the test dataset, while RF and GD-SVM were tested three times with different random seeds.

For **deep learning** models, we used 10% of the training data as the development dataset and choose the best hyperparameters based on the development performance (weighted F1). We found the best hyperparameters for RoBERTa and RoBERTa-CON using the grid search. We considered two hyperparameters, i.e., batch size and learning rate. We search the batch size in a set of {*4, 8, 16*}; and the learning rate in a set of {*5e-5, 3e-5, 2e-5, 1e-5, 5e-6, 2e-6*}, with 20 epochs for EMO-COG and 50 epochs for all multilabel tasks. We conducted hyperparameter optimization for two types of tasks, COG-8 and EMO-COG. Due to the high computational costs of the fine-tuning process, the best combination for COG-8 was used for other multilabel tasks, including EMO-8, COG-FULL and EMO-FULL.

**Table 1.** Best hyperparameters for RoBERTa and RoBERTa-CON used in our experiments

|  | RoBERTa | | | RoBERTa-CON | | |
|---|---|---|---|---|---|---|
|  | Learning rate | Batch size | Best epoch | Learning rate | Batch size | Best epoch |
| EMO-COG | 1e-5 | 4 | 8 | 5e-6 | 8 | 3 |
| EMO-8 | 3e-5 | 8 | 12 | 1e-5 | 8 | 19 |
| COG-8 | 3e-5 | 8 | 12 | 1e-5 | 8 | 19 |
| EMO-FULL | 3e-5 | 8 | 10 | 1e-5 | 8 | 11 |
| COG-FULL | 3e-5 | 8 | 8 | 1e-5 | 8 | 45 |

After getting the best hyperparameters, we trained RoBERTa and RoBERTa-CON on the training set for *three times*. Each time we randomly sampled 10% of the training data as the development set and recorded the development performance of each epoch. We averaged the results of each epoch over three runs and chose the best epoch based on the average development performance (weighted F1 score). For EMO-COG task, we set the maximum number of epochs to 20, and for the other four tasks to 50. Finally, we used the best hyperparameter set to train the models on the whole training set respectively and tested them on the testing set. We summarized the best hyperparameters in Table 1.

### 3.4 Evaluation Metrics and Baseline

We calculated three metrics: macro F1 score (M-F1), weighted F1 score (W-F1) and accuracy (the latter was replaced by hamming loss for multilabel classification tasks) [13]. M-F1 returns the average of F1 scores across all classes with equal weight. W-F1 computes the average using weights that reflect the proportion of each class in the dataset. M-F1 and W-F1 scores can be calculated using the equations (1) and (2) respectively:

$$Macro\ F1 = \frac{1}{N}\sum_{i=0}^{N} F1_i, \tag{1}$$

$$Weighted\ F1 = \sum_{i=0}^{N} W_i F1_i, \tag{2}$$

where $N$ is the number of classes, and $F1_i$ is the F1 score of the class $i$, and $W_i$ is the weight, assigned to class $i$; $W_i = \frac{number\ of\ instances\ of\ class\ i}{training\ set\ size}$. The general formula for F1 is represented by the formula (3):

$$F1 = \frac{TP}{TP + \frac{1}{2}(FP + TP)}, \tag{3}$$

where $TP$ is true positives (correctly labeled as belonging to the class) and $FP$ is the false positives (incorrectly labeled as belonging to the class).

Accuracy (ACC) in machine learning refers to the ratio of correctly predicted items to the total input size, and can be estimated using the following equation:

$$ACC = \frac{number\ of\ correct\ predictions}{total\ number\ of\ predictions}, \tag{4}$$

Since in multilabel classification a prediction can be fully correct, partially correct (some labels were predicted correctly, some were not) or fully incorrect, it is not possible to calculate accuracy using the standard formula. For that reason, we used Hamming loss (HL) instead of accuracy for multilabel classification tasks. HL describes the fraction of incorrectly predicted items and is represented by the formula (5). As this is the loss function, higher scores correspond to less accurate models.

$$HL = \frac{1}{|N||L|}\sum_{1}^{|N|}\sum_{1}^{|L|} xor(y_{ij}, z_{ij}), \tag{5}$$

where $y_{ij}$ is the target, $z_{ij}$ is the prediction, and $xor()$ is the exclusive "or" operator, i.e., $xor(y_{ij}, z_{ij})$ is equal to 0, when target and prediction are equal, and to 1 otherwise. In the next section, we present the M-F1, W-F1, ACC and HL in the percentage form and use W-F1 as our main metric.

**Baseline.** To investigate how much models can learn the task, we constructed a simple baseline in which all the utterances were classified as the dominant class (e.g., non-emotion in EMO-COG classification).

## 4 Results

### 4.1 EMO-COG Classification

The results of the EMO-COG classification (emotion vs. non-emotion) are provided in the Table 2. Among the conventional machine learning methods, GD-SVM demonstrated the best performance by every metric, achieving the average W-F1 of 88.3. The improvement in M-F1 over the baseline was around 32.61, which demonstrates the model's ability to identify the less represented emotional utterances. Our deep learning model, RoBERTa-CON, achieved the best performance with W-F1 of 98.71.

Table 2. EMO-COG classification results

|  | W-F1 | M-F1 | ACC |
|---|---|---|---|
| Conventional machine learning | | | |
| Baseline | 74.75 | 45.24 | 82.61 |
| Naïve Bayes | 66.40 | 53.37 | 61.93 |
| AdaBoost + Naïve Bayes | 72.03 | 57.74 | 69.09 |
| Random Forest | 84.87 ± 0.41 | 69.68 ± 0.95 | 84.25 ± 5.83 |
| GD-SVM | **88.30 ± 0.15** | **77.85 ± 4.53** | **89.43 ± 0.11** |
| Logistic Regression | 87.12 | 74.93 | 88.86 |
| Deep learning models | | | |
| RoBERTa | 90.31 ± 0.47 | 79.54 ± 2.71 | 92.92 ± 0.92 |
| RoBERTa-CON | <u>**98.71 ± 0.13**</u> | <u>**98.08 ± 0.21**</u> | <u>**98.38 ± 0.13**</u> |

Note. **Bold** font indicates the best performance in each group of methods; <u>underscores</u> indicate the best performance across all the models.

### 4.2 Multilabel Classification

The results of the multilabel classification for the coarse-grained labels (EMO-8 and COG-8) are provided in the Table 3. The results of EMO-FULL and COG-FULL are presented in the Table 4.

**Conventional Machine Learning**. Similar to the previous task, GD-SVM demonstrated the best performance among conventional models for EMO-8 and COG-8 tasks. While the absolute scores are lower, the improvement over the baseline clearly shows that the models are able to distinguish between classes. W-F1 of GD-SVM are improved over 37.15 points respectively for EMO-8 compared to the baseline; for COG-8 the improvement is slightly smaller with 27.3 points for W-F1. For EMO-FULL, the best W-F1 is still reached by GD-SVM, but NB achieved the best performance on COG-FULL.

**Deep Learning**. For the most multiclass-multilabel tasks, our RoBERTa-CON again outperformed RoBERTa and all conventional machine learning methods, except for the EMO-FULL task. Our RoBERTa-CON acquired W-F1 of 71.01, 68.31, 45.81 on EMO-8, COG-8, COG-FULL respectively. RoBERTa achieved the best W-F1 of 60.40 on EMO-FULL. Due to a large number of labels and data imbalance, the models however performed poor for minor classes (see error analysis in Section 4.3). As the difficulty of tasks increased (from coarse-grained to fine-grained labels), we noticed that the model performance decreased, including the best performed model RoBERTa-CON. There is still a large gap needed to fill for fine-grained classifications of psychological data by using machine learning methods. We will improve this in the future work.

Table 3. EMO-8 and COG-8 classification results

|  | EMO-8 | | | COG-8 | | |
|---|---|---|---|---|---|---|
|  | W-F1 | M-F1 | HL | W-F1 | M-F1 | HL |
| Conventional machine learning | | | | | | |
| Baseline | 11.45 | 5.38 | 18.55 | 16.14 | 6.96 | 19.51 |
| NB | 31.96 | 23.97 | 22.06 | 40.16 | 23.85 | 32.78 |
| AdaBoost + NB | 31.22 | 21.34 | 19.85 | 35.42 | 22.32 | 26.26 |
| RF | 40.31 ± 0.61 | 25.07 ± 0.83 | **10.24 ± 0.12** | 25.05 ± 0.15 | 12.50 ± 0.13 | 17.41 ± 0.02 |
| GD-SVM | **48.60 ± 2.33** | **37.42 ± 4.53** | 12.31 ± 0.52 | 43.44 ± 1.50 | 24.44 ± 0.35 | 17.04 ± 0.34 |
| LR | 39.93 | 24.73 | 12.83 | 42.69 | 24.42 | **16.32** |
| Deep learning models | | | | | | |
| RoBERTa | 69.08 ± 4.92 | 60.48 ± 7.78 | 7.76 ± 0.78 | 65.45 ± 0.49 | 49.48 ± 3.88 | 11.18 ± 0.43 |
| RoBERTa-CON | **71.01 ± 1.84** | **62.21 ± 1.99** | **7.30 ± 0.53** | **68.31 ± 0.39** | **51.76 ± 0.65** | **10.38 ± 0.18** |

Table 4. EMO-FULL and COG-FULL classification results

|  | EMO-FULL | | | COG-FULL | | |
|---|---|---|---|---|---|---|
|  | W-F1 | M-F1 | HL | W-F1 | M-F1 | HL |
| Conventional machine learning | | | | | | |
| Baseline | 8.66 | 1.91 | 7.83 | 4.11 | 0.70 | 4.62 |
| NB | 16.97 | 7.87 | 9.94 | **15.09** | **6.80** | 15.21 |
| AdaBoost + NB | 21.06 | 9.87 | 9.33 | 14.66 | 6.78 | 11.03 |
| RF | 24.74 ± 0.57 | 10.46 ± 0.88 | **4.69 ± 0.13** | 4.56 ± 1.07 | 2.31 ± 0.55 | 4.39 ± 0.02 |
| GD-SVM | **40.07 ± 0.85** | **23.93 ± 3.01** | 5.76 ± 0.17 | 9.97 ± 1.06 | 3.96 ± 0.30 | 4.34 ± 0.05 |
| LR | 24.41 | 9.42 | 4.78 | 7.76 | 2.82 | **4.06** |
| Deep learning models | | | | | | |
| RoBERTa | **60.40 ± 0.18** | **48.01 ± 4.65** | 3.97 ± 0.16 | 44.60 ± 0.77 | 29.62 ± 2.76 | 4.13 ± 0.08 |
| RoBERTa-CON | 51.92 ± 1.69 | 27.66 ± 3.02 | **3.66 ± 0.16** | **45.81 ± 0.13** | **30.26 ± 0.76** | **4.03 ± 0.03** |

### 4.3 Error Analysis

We conducted an error analysis and inspected the best model's (our RoBERTa-CON) performance for each individual class. As was mentioned previously, all models demonstrated poorer prediction accuracy for the non-emotion in the EMO-COG task which can be explained by the data imbalance. Figure 2 visualizes the classification results of RoBERTa-CON model on the EMO-8 and COG-8 tasks. The visualization reveals that the best performance was achieved for labels such as 'joy' and 'anticipation' for EMO-8, and 'agreement' and 'description' for COG-8. Those labels correspond to the dominant classes. The model showed poorer performance for minor classes, such as 'fear' (6% of the EMO-8 dataset) and 'suggestion' (7% of the COG-8 dataset).

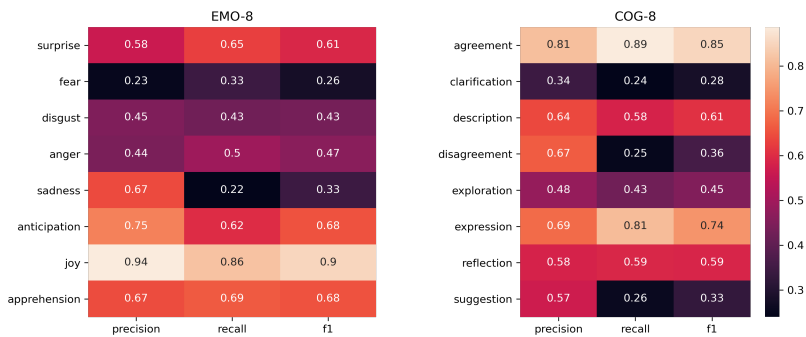

**Fig. 2.** The classification results for RoBERTa-CON model for EMO-8 (left) and COG-8 (right)

For fine-grained classifications, some less represented labels were not recognized by RoBERTa-CON, e.g., the label 'express fear' for EMO-FULL or 'advise' and 'disapprove' for COG-FULL. It is worth mentioning that for some labels the lack of representation in the training dataset does not affect the model's ability. For example, labels 'pause' and 'unintelligible response' from COG-FULL have the highest precisions among others in the same task.

## 5    Conclusion and Discussion

In this study, we investigated the utility of conventional machine learning as well as deep learning methods for the automated labeling of utterances in conversations on adulthood transition. To handle three severe data challenges, we developed a hierarchical classification system to tackle our problems from coarse-grained level to fine-grained level. Our results suggested that pretrained deep learning models outperform conventional methods. To further tackle these challenges, we performed a domain adaptation for the original pretrained RoBERTa on large-scale counseling data. Our adapted model, RoBERTa-CON, was almost comparable to the human coder demonstrating W-F1 as high as 98% on EMO-COG task and achieved the best performance on most multiclass-multilabel tasks. Although coarse-grained labels are not equivalent to the fine-grained labels, they can serve as a mid-step to help researchers understand their data.

Some challenges were still persistent. The model performance was substantially decreasing as the number of labels increased. We also noticed that the data imbalance affected the individual performance for minor labels. We plan to improve those limitations in future work by using transfer learning, semi-supervised learning, etc.